\documentclass[conference]{IEEEtran}
\usepackage[T1]{fontenc}
\usepackage[utf8]{inputenc}
\usepackage{color,array}
\usepackage{float}
\usepackage{times}
\usepackage{soul}
\usepackage{url}
\usepackage{multirow}
\usepackage{booktabs}
\usepackage{comment}

\usepackage{cite}
\usepackage{amsmath,amssymb,amsfonts}
\usepackage{graphicx}
\usepackage{textcomp}
\usepackage{xcolor}
\usepackage{algorithm}
\usepackage[noend]{algpseudocode}
\def\BibTeX{{\rm B\kern-.05em{\sc i\kern-.025em b}\kern-.08em
    T\kern-.1667em\lower.7ex\hbox{E}\kern-.125emX}}
\begin{document}

\title{Client Contribution Normalization for Enhanced Federated Learning}

\author{
    \IEEEauthorblockN{Mayank Kumar Kundalwal\IEEEauthorrefmark{1}, Anurag Saraswat\IEEEauthorrefmark{2}, Ishan Mishra\IEEEauthorrefmark{1}, and Deepak Mishra\IEEEauthorrefmark{1}}
    \IEEEauthorblockA{\IEEEauthorrefmark{1}Indian Institute of Technology Jodhpur, Jodhpur, India\\
    \IEEEauthorrefmark{1}Email: kundalwal.1@iitj.ac.in, mishra.10@iitj.ac.in, dmishra@iitj.ac.in}
    \IEEEauthorblockA{\IEEEauthorrefmark{2}Teradata India Private Limited, Telangana, India\\
    \IEEEauthorrefmark{2}Email: anurag.saraswat@teradata.com}
}

\maketitle

\begin{abstract}
Mobile devices, including smartphones and laptops, generate decentralized and heterogeneous data, presenting significant challenges for traditional centralized machine learning models due to substantial communication costs and privacy risks. Federated Learning (FL) offers a promising alternative by enabling collaborative training of a global model across decentralized devices without data sharing. However, FL faces challenges due to statistical heterogeneity among clients, where non-independent and identically distributed (non-IID) data impedes model convergence and performance. This paper focuses on data-dependent heterogeneity in FL and proposes a novel approach leveraging mean latent representations extracted from locally trained models. The proposed method normalizes client contributions based on these representations, allowing the central server to estimate and adjust for heterogeneity during aggregation. This normalization enhances the global model's generalization and mitigates the limitations of conventional federated averaging methods. The main contributions include introducing a normalization scheme using mean latent representations to handle statistical heterogeneity in FL, demonstrating the seamless integration with existing FL algorithms to improve performance in non-IID settings, and validating the approach through extensive experiments on diverse datasets. Results show significant improvements in model accuracy and consistency across skewed distributions. Our experiments with six FL schemes—FedAvg, FedProx, FedBABU, FedNova, SCAFFOLD, and SGDM highlight the robustness of our approach. This research advances FL by providing a practical and computationally efficient solution for statistical heterogeneity, contributing to the development of more reliable and generalized machine learning models.
\end{abstract}


\section{Introduction}
Mobile devices have become ubiquitous sources of decentralized and heterogeneous data. These devices, owned and operated by individuals generate vast amounts of diverse data. Machine learning models predominantly rely on centralized training, where data from these devices is collected on a single central server. However, this approach entails transferring data to a centralized server, incurring substantial communication costs and posing risks to data privacy \cite{poushter2016smartphone}. Moreover, it raises significant concerns regarding data privacy and security \cite{goyal2017accurate}. To address these challenges, the concept of isolated training at the user level has been proposed. While this approach preserves data privacy by keeping data on the user's device, it encounters limitations related to the sufficiency of individual user data for training reliable machine learning models.

Federated Learning (FL) \cite{konevcny2016federated,shokri2015privacy,mcmahan2017communication} emerges as a subset of collaborative learning specifically designed to address privacy concerns inherent in centralized learning. FL operates within a distributed machine learning framework, enabling models to be trained on local data without the necessity of sharing the data itself. Despite its potential, FL encounters its own array of challenges. One primary concern is statistical heterogeneity among clients. In real-world scenarios, data distributions across clients are often non-independent and identically distributed (non-IID), leading to challenges in model convergence and performance. Federated averaging (FedAvg), a widely used aggregation method in FL, faces significant drawbacks when dealing with non-IID data distributions. These drawbacks include a biased global model, slower convergence, poor generalization, and client drift.

This paper investigates challenges associated with data dependency in FL, focusing on addressing statistical heterogeneity. Our proposed approach utilizes mean latent representations extracted from locally trained models. The key strength of our method lies in normalizing client contribution values through their respective mean latent representations. The server evaluates distances among clients in the representation space, enabling the estimation of heterogeneity and subsequent adjustment of each client's model contribution during the aggregation process. This normalization enhances the generalization of the global model. Unlike existing methods that primarily focus on adjusting local training processes or introducing regularization terms, our approach operates at the aggregation level. This allows for seamless integration with existing FL algorithms without requiring modifications to local training procedures or model architectures, making it a more versatile and practical solution for addressing non-IID data challenges in FL.

The main contributions of this work are as follows:
\begin{itemize}
    \item We propose a novel normalization scheme using mean latent representations to address statistical heterogeneity in FL.
    \item We demonstrate the seamless integration of our method with existing FL algorithms, enhancing their performance in non-IID settings.
    \item Through extensive experiments, we show significant improvements in model accuracy and consistency across diverse datasets and skewed distributions.
\end{itemize}

To evaluate the effectiveness of our proposed approach, we conduct a comprehensive series of experiments across various datasets with differing data distributions. We integrate our approach with six different FL schemes: FedAvg \cite{bonawitz2017practical}, FedProx \cite{li2018federated}, FedBABU \cite{oh2022fedbabu}, FedNova \cite{wang2020tackling}, SCAFFOLD \cite{karimireddy2020scaffold}, and SGDM \cite{liu2020accelerating}. We consistently observe enhancements in the performance of the global model, validating the efficacy of our approach.

The subsequent sections of this paper are structured as follows. Section 2 reviews related literature, Section 3 discusses the problem setting and our approach, Section 4 presents experimental results. Finally, Section 6 presents the conclusion of our paper.

\section{Related Work}
FL is a continuously evolving field in machine learning with several possibilities. It holds a crucial role in distributed machine learning, particularly when addressing data-dependent challenges. One of the central issues in FL is handling non-IID data, which presents a substantial challenge. Addressing this challenge is crucial for improving the robustness and generalization of FL models across diverse and heterogeneous data sources. This section provides a summary of the existing work in the field and clarifies the motivation behind our research.

Conventional FL methods rely on various techniques, such as Federated Stochastic Gradient Descent (FedSGD) and Federated Averaging (FedAvg), for global model aggregation. FedSGD involves the selection of multiple clients to compute gradients based on their local data, which are subsequently averaged to generate a global model update. FedAvg allows clients to perform multiple batch updates and share model weights with the centralized server, rather than gradients. In FedAvg, the centralized server aggregates the received model weights using a weighted average \cite{bonawitz2017practical}.

System heterogeneity is also a significant challenge in FL, as participating clients may have varying storage, communication, and computation capabilities. Furthermore, clients may drop out during training due to factors such as energy constraints and connectivity issues \cite{bonawitz2019towards}. Communication cost poses yet another major challenge in FL, often serving as a bottleneck when model convergence requires numerous rounds, each involving the transfer of large model parameters. Some works address this by employing model compression techniques such as weight pruning and quantization, though these methods can result in performance loss due to the inherent lossy compression \cite{han2015deep,9407898}. Other techniques like variance reduction \cite{liang2019variance}, dynamic regularizers, and the use of different client subsets for distinct communication rounds have been introduced to reduce communication costs and handle client heterogeneity. FedPCC represents an efficient methodology based on the parallelism of communication and computation among devices for FL in wireless networks \cite{zhang2022fedpcc}.

\subsection{Statistical Heterogeneity in Federated Learning}
There exists a vast amount of literature in deep learning to handle statistical heterogeneity using approaches like multitask learning and meta-learning. Recently, these methods of handling non-IID data have been incorporated in FL settings \cite{marfoq2021federated,pang2020realizing,chen2020zero}. For instance, MOCHA allows personalization in FL by learning separate but related models at each participating device using multitask learning \cite{smith2017federated}. Despite these advances, the key challenge of statistical heterogeneity still remains open. Hsu \textit{et al.} explore the performance of FL approaches on classification tasks and show that FedAvg starts performing poorly with increasing non-IIDness \cite{hsu2020federated}. Other works focus on heterogeneity in computation cost in FL \cite{tran2019federated,bonawitz2019towards}. Additionally, Luo \textit{et al.} introduced the EdgeFed (Edge FL) approach, which employs an edge server to aggregate weight updates between the centralized server and the clients \cite{luo2021cost}.

Although various methods have targeted data heterogeneity in FL, most of them introduce additional regularizers to handle the non-IID data at the client end. FedProx uses a dynamic regularizer to handle non-IID data at the client end \cite{li2018federated}. SCAFFOLD highlights the drawbacks of FedAvg and proposes the use of control variate (variance reduction) to handle client drift in the presence of heterogeneous data, \cite{karimireddy2020scaffold}. The recently introduced FedBABU algorithm uniquely updates the model's body (extractor) during federated training, leaving the head unaltered and randomly initialized \cite{oh2022fedbabu}. FedNova addresses heterogeneity by normalizing the number of local iterations at each client, assigning weights to clients during aggregation based on these normalized iterations \cite{wang2020tackling}. In FL, Stochastic Gradient Descent with Momentum (SGDM) \cite{zinkevich2010parallelized} is employed to optimize model parameters across distributed clients.

In FL, the contribution factor plays a pivotal role in the aggregation process, influencing how individual models contribute to the global model. FedAvg utilizes the number of samples from each local model as a key factor in determining its contribution value during the aggregation process \cite{bonawitz2017practical}. Various similarity-based approaches have been proposed to tackle issues posed by non-IID data and malicious clients. The adaptive federated averaging method, proposed by Muñoz-González \textit{et al.}, calculates the cosine similarity between local model gradients and the weighted average of all local models in each round and aggregates the global model using contribution values \cite{munoz2019byzantine}. Similarly, the FLTrust approach introduced by Cao \textit{et al.} measures cosine similarity between a local model and the server model trained with a small root dataset \cite{cao2020fltrust}.

While these approaches have made significant progress, several limitations persist. Many methods rely on specific model architectures or introduce substantial computational overhead. Some require modifications to local training procedures or model structures, limiting their applicability. The trade-off between addressing heterogeneity and maintaining model generalization remains a challenge.

Our work addresses these gaps by proposing a flexible, computationally efficient plug-in solution that integrates seamlessly with existing FL algorithms. By utilizing mean latent representations, we offer a versatile approach to normalize client contributions without modifying local training procedures or model architectures. This not only enhances the performance of FL algorithms in non-IID settings but also provides a practical framework for real-world applications where data heterogeneity is prevalent. Through extensive experiments, we demonstrate significant improvements in model accuracy and consistency across diverse datasets and skewed distributions.

\begin{figure*}[t]
\begin{minipage}[t]{0.48\linewidth}
    \includegraphics[width=\linewidth]{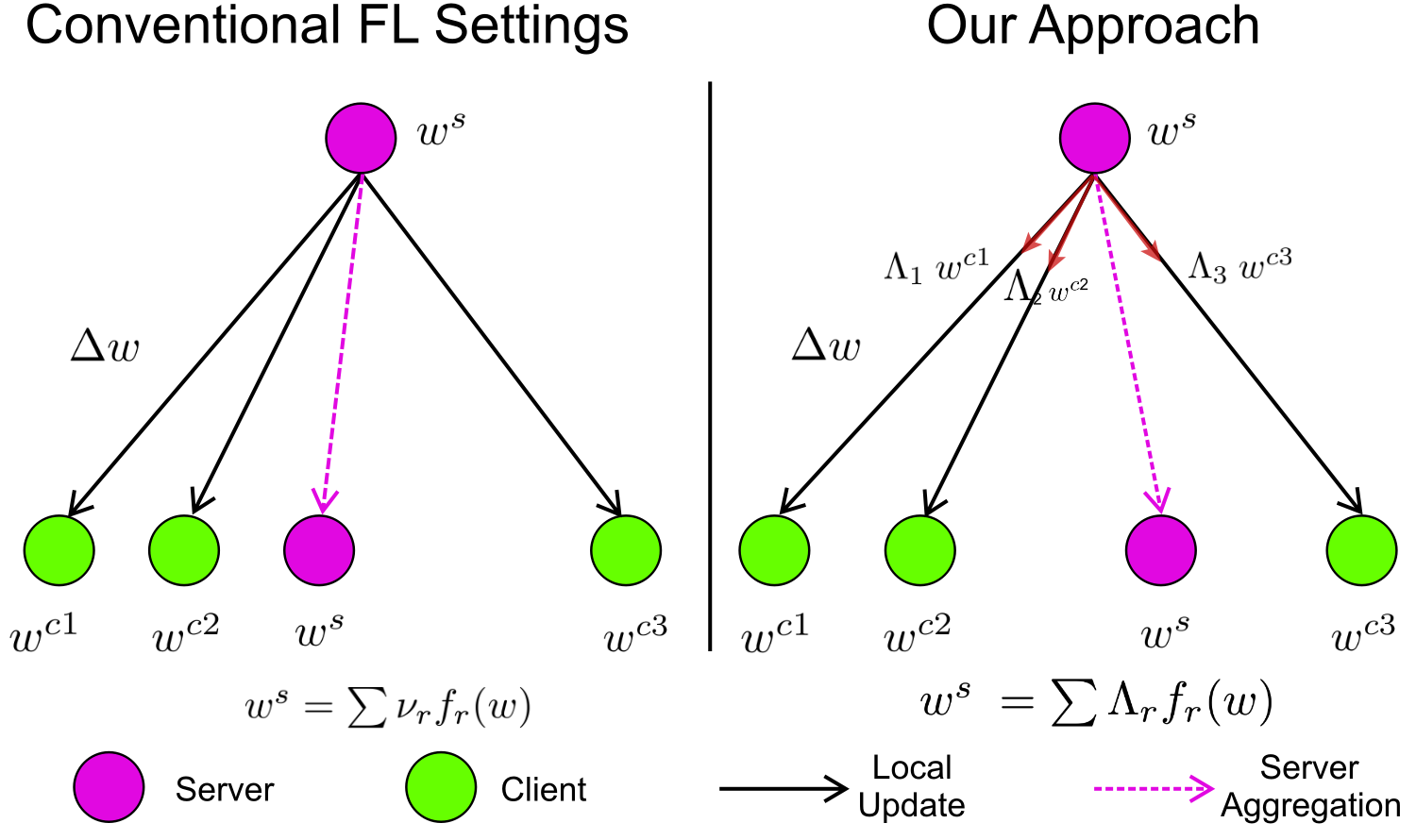}
\end{minipage} %
    \hfill%
\begin{minipage}[t]{0.48\linewidth}
    \includegraphics[width=\linewidth]{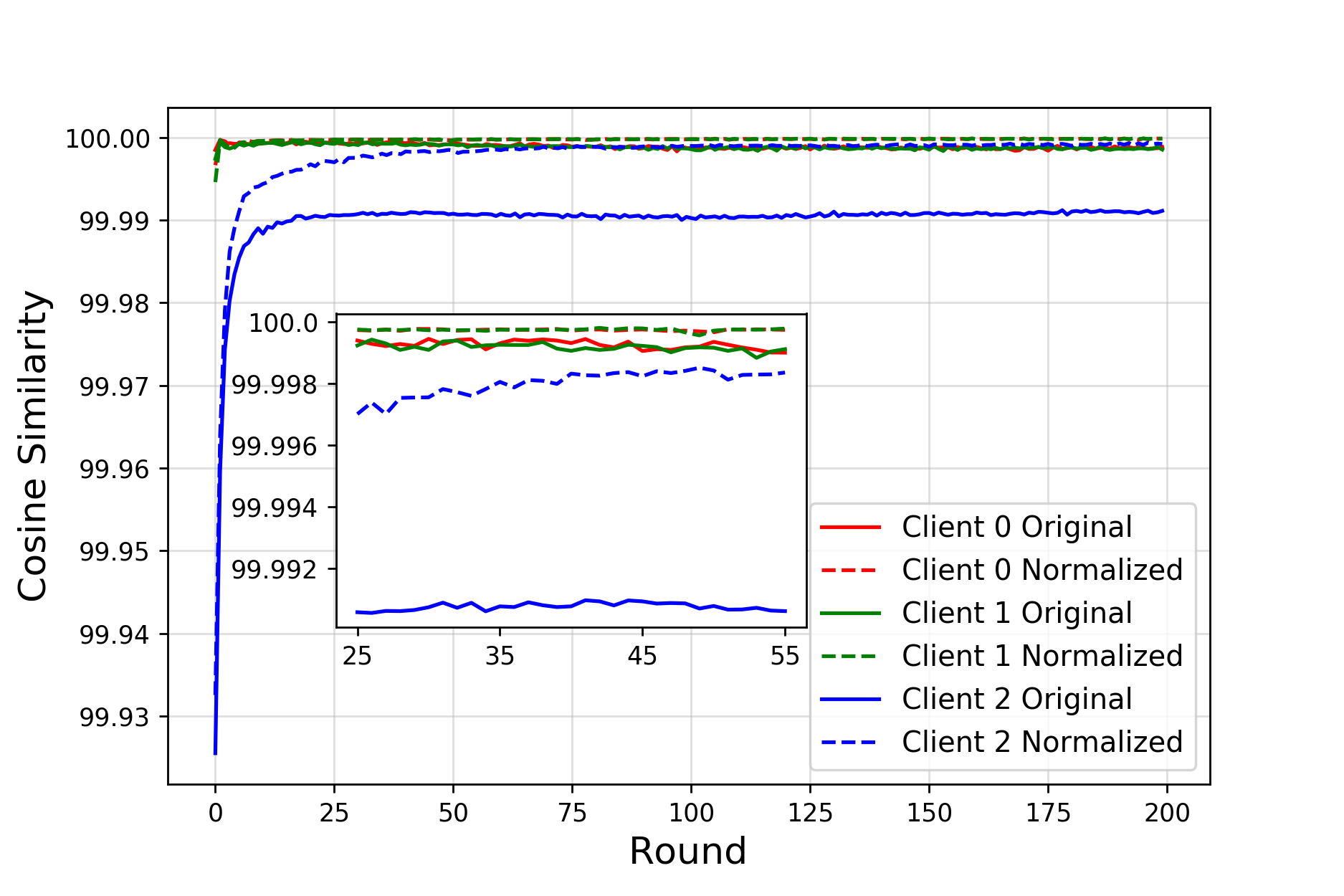}
\end{minipage}
\begin{minipage}{0.49\linewidth}
    \centering
    (a)
    \label{a}
\end{minipage}
\begin{minipage}{0.49\linewidth}
    \centering
    (b)
    \label{b}
\end{minipage}
\caption{Illustration of our proposed approach. Clients 1 and 2 have similar data distributions, while client 3 has a different data distribution. As a result, the locally trained model of client 3, denoted as $w^{c3}$, will be far from the models of the other two clients, $w^{c1}$ and $w^{c2}$, which are expected to be similar due to their similar data distributions. In the conventional FL setting, the same weight is assigned to all three clients during model aggregation, assuming that they have trained for the same number of epochs. However, in our proposed approach, the weight assigned to each client is influenced by the distance between that client's model and the models of other clients.}
\label{fig:my_label}
\end{figure*}

\section{Methodology}
In FL, the central server aggregates the local models received from participating clients and broadcasts the aggregated model to all clients. In the conventional FL setting~\cite{bonawitz2017practical}, the overall objective function $f(w)$ is defined as the weighted sum of the local objectives $f_r(w)$:
\begin{equation}
     f(w^*) = \min_{w} f(w) = \sum^{R}_{r=1} \nu_{r}f_r(w)
\end{equation}
Here, $f(w^*)$ gives the aggregated loss after convergence, $w^*$ represents the optimal weights, $R$ is the total number of clients and $\nu_r$ is the weight associated with client $r$, ensuring that $\sum^{R}_{r=1}\nu_r = 1$. The local objective function $f_r(w)$ is defined as:
\begin{equation}
f_{r}(w) = \mathbb{E}_{x_r \sim P_{D_r}} [l_{r}(w;x_{r})]
\label{eq2}
\end{equation}
The expected value $\mathbb{E}_{x_r \sim D_r}$ is taken over all possible data points $x_r$ drawn from the dataset $D_r$ according to the distribution $P_{D_r}$.

In FedAvg, the importance factor depends on the number of samples of that particular client. However, different data distributions at clients induce various data-dependent issues, and information about only the number of samples at each client is not enough to account for these issues \cite{dhada2020empirical}.

\begin{algorithm}[t]
\caption{PROPOSED\_APPROACH}
\begin{algorithmic}[1] 
\State {\textbf{Server Side Execution:}}
\State Broadcast initialized model $w_0^S$ 
\State $R \leftarrow $(random group of clients)
\For{each communication round $c = 0, 1, 2.....C$}
\For{each client in $\textit{r $\in$ R}$ \textbf{in parallel}}
\State $w^{r}_{c},z_r \leftarrow$ Client$(r,w_{c}^{S} )$

\EndFor
\State Compute $\boldsymbol{\Lambda}$, $\boldsymbol{\nu}$
\State $w_{c+1}^{S} \leftarrow$ $\sum^{R}_{r=1}\frac{( \Lambda_r \nu_r )}{\mathbf{\Lambda} \cdot \boldsymbol{\nu} }w^{r}_{c}$
\EndFor
\Function{Client}{r,w}
    \State receives broadcasted model $w$
    \State $w \leftarrow w_0$
    \For{each client in $\textit{r $\in$ R}$ \textbf{in parallel}}
        \State Compute mean representation $z_r$
        \For{each epoch $i = 0, 1, 2, \dots $}
            \State $w^{r}_{c} \leftarrow {argmin}_w f_r(w)$
            \If{last epoch}
            \State Compute $z_r$
            \EndIf
        \EndFor
    \EndFor
    \State return $w^{r}_{c}$, $z_r$
\EndFunction

\end{algorithmic}
\label{alg1}
\end{algorithm}

\subsection{Mean Latent Representation}
At each client, we utilize the activations derived from the final hidden layer of the local model as a representation of semantic features, capturing valuable information about the variation in data across different clients. Each client transmits both its model weights $w_r$ and mean latent representation $z_r$ to the centralized server. During aggregation, the server calculates similarity among the $z_r$ values of all participating clients using the Cosine similarity metric:
\begin{equation}
    \cos(z_r,z_p) = \frac{\langle z_{r},z_{p}\rangle}{||z_{r}||\cdot||z_{p}||}
\end{equation}
The similarity matrix $S$ is defined as follows:
\begin{equation}
  S(r,p)=\begin{cases}
    \cos(z_r,z_p), & \text{if $r \neq p$}\\
    1, & \text{if $r = p$}
  \end{cases}
\end{equation}
The contribution factor $\Lambda_r$ for client $r$ is defined as:

\begin{equation}
    \Lambda_r = \frac{\sum_{q \neq r}exp(\sum_{p}S(q,p))}{\sum_{q}exp(\sum_{p}S(q,p))}
    \label{att_eq1}
\end{equation}

\subsection{Integration with Existing FL Algorithms}
The proposed approach handles variability in data distribution among participating clients using $\boldsymbol{\Lambda} = [\Lambda_1 \hspace{0.1cm} \Lambda_2 \hspace{0.1cm} \Lambda_3 \hspace{0.1cm} \dots \hspace{0.1cm}\Lambda_R]^{T} $. It normalizes the contribution of individual clients during aggregation by integrating $\boldsymbol{\Lambda}$ into the mainstream FL framework and modifying the global objective as follows:

\begin{equation}
    \min_{w} f(w) = \sum^{R}_{r=1}\frac{( \Lambda_r \nu_r )}{\mathbf{\Lambda} \cdot \boldsymbol{\nu} }f_r(w)
\end{equation}
where $\nu_r$ is the importance function defined by the mainstream FL framework, and $\mathbf{\Lambda} \cdot \boldsymbol{\nu}$ is the dot product of the importance function vector ($\boldsymbol{\nu}$) and the proposed $\boldsymbol{\Lambda}$.

This integration allows our approach to be easily incorporated into existing FL algorithms. For example, in FedAvg, we replace the standard aggregation step with our normalized version:

\begin{equation}
    w_{t+1} = \sum^{R}_{r=1}\frac{( \Lambda_r \nu_r )}{\mathbf{\Lambda} \cdot \boldsymbol{\nu} }w^r_t
\end{equation}

where $w_{t+1}$ is the global model at round $t+1$, and $w^r_t$ is the local model of client $r$ at round $t$.

Fig. \ref{fig:my_label} (a) illustrates the behavior of our proposed approach, controlled by the contribution factor, in a simulated data setting with three clients. Two clients have data from eight CINIC-10 classes, while the third has data from the remaining two. We compare FedAvg and Normalized FedAvg by plotting the similarity between the local and global weight matrices after each round, as shown in Fig. \ref{fig:my_label}(b). The proposed normalization approach prioritizes client 3, bringing its model closer to the global model, unlike FedAvg, which treats all clients equally, resulting in client 3's model being further from the global model. The inset highlights this effect, demonstrating the effectiveness of our approach in achieving a more balanced weight distribution among clients. Algorithm \ref{alg1} outlines the methodology employed in the proposed approach.

\subsection{Computational Overhead}
Our approach introduces three main computational steps: (1) mean latent representation computation, (2) inter-client similarity calculation, and (3) contribution factor derivation. These processes are efficiently integrated into the FL pipeline, occurring once per communication round without requiring significant additional data transfer.

\subsection{Normalization with Temperature Scaling}
To reduce abrupt changes in contribution values across consecutive rounds, we use temperature scaling. We modify equation \eqref{att_eq1} as:

\begin{equation}
    \Lambda_{r_T} = \frac{\sum_{q \neq r}exp(\sum_{p}S(q,p)/T)}{\sum_{q}exp(\sum_{p}S(q,p)/T)}
    \label{att_eq}
\end{equation}

where $\Lambda_{r_T}$ is the normalized contribution factor for node $r$ with temperature scaling. A temperature value less than 1 results in refinement of the contribution value obtained using the mean latent representation. We perform all our experiments using this temperature scaling approach to ensure consistent and stable evaluation of client contributions.

\section{Experimental Results}
This section presents a rigorous analysis of our proposed approach's performance across diverse data settings, comparing it with baseline methods. We employ three distinct datasets to comprehensively assess the robustness and efficacy of our method under varying degrees of data heterogeneity.

\subsection{Implementation Details}
We utilize a VGG-16 model architecture for all experiments. Each client executes two local epochs per communication round, with a total of 200 communication rounds. The Adam optimizer is employed with a batch size of 64. The mean latent representation is derived from the final hidden layer of VGG-16, projecting the output into a 128-dimensional vector. Our comparative analysis involves six baseline FL techniques: FedAvg \cite{mcmahan2017communication}, FedProx \cite{li2018federated}, FedBABU \cite{oh2022fedbabu}, SCAFFOLD \cite{karimireddy2020scaffold}, SGDM \cite{zinkevich2010parallelized}, and FedNova \cite{wang2020tackling}.

\subsection{Datasets and Data Distribution}
We conduct experiments on CIFAR-10 \cite{krizhevsky2009learning}, FEMNIST \cite{cohen2017emnist}, and CINIC-10 \cite{darlow2018cinic}. CIFAR-10 is used to evaluate performance under controlled non-IID settings with varying degrees of heterogeneity. FEMNIST is selected to assess the approach's efficacy on naturally heterogeneous data with imbalanced class distributions. CINIC-10 is employed to test robustness against augmented and transformed data distributions. For CIFAR-10 and CINIC-10, we generate non-IID distributions using a Dirichlet distribution with concentration parameter $\alpha$. Lower $\alpha$ values induce higher data heterogeneity, allowing us to systematically evaluate our method's performance across a spectrum of non-IID scenarios.

\subsection{Results and Analysis}

\subsubsection{CIFAR-10}
Table \ref{table_cifar} presents the results on CIFAR-10 with 50 clients and $\alpha = 0.1$ and $0.5$. The proposed normalization approach consistently improves performance across all baseline methods, with gains of 3-5\% in accuracy. SGDM with normalization achieves the highest accuracy (82.68\% for $\alpha = 0.5$), indicating the synergistic effect of momentum and our contribution-based normalization. The performance gap between normalized and original methods is more pronounced for $\alpha = 0.1$, suggesting our approach's enhanced efficacy in highly heterogeneous settings. Notably, FedAvg shows the lowest performance among all methods, particularly in the highly non-IID scenario ($\alpha = 0.1$), highlighting its limitations in handling heterogeneous data distributions.


\begin{table}[t]
\setlength{\tabcolsep}{5pt}
\centering
\caption{Comparison of experimental results for different FL schemes integrated with the normalization approach on the CIFAR-10 dataset. The numbers in brackets indicate the increase in accuracy from the original to the normalized approach.}
\begin{tabular}{ccccc}
\toprule
\textbf{Approach} & \multicolumn{2}{c}{\textbf{Original}} & \multicolumn{2}{c}{\textbf{Normalized}} \\ \midrule
 & $\alpha = 0.1$ & $\alpha = 0.5$ & $\alpha = 0.1$ & $\alpha = 0.5$ \\ \midrule
\textbf{FedAvg}   & 69.68 & 74.36 & 74.44\textsubscript{\scriptsize{(4.76)}} & 79.29\textsubscript{\scriptsize{(4.93)}} \\ 
\textbf{FedProx}  & 72.95 & 75.18 & 77.04\textsubscript{\scriptsize{(4.09)}} & 80.32\textsubscript{\scriptsize{(5.14)}} \\ 
\textbf{SCAFFOLD} & 71.96 & 75.26 & 77.98\textsubscript{\scriptsize{(6.02)}} & 81.38\textsubscript{\scriptsize{(6.12)}} \\ 
\textbf{SGDM}     & 72.10 & 78.14 & 76.02\textsubscript{\scriptsize{(3.92)}} & 82.68\textsubscript{\scriptsize{(4.54)}} \\ 
\textbf{FedNova}  & 71.02 & 76.26 & 76.88\textsubscript{\scriptsize{(5.86)}} & 81.58\textsubscript{\scriptsize{(5.32)}} \\ 
\textbf{FedBABU}  & 72.92 & 75.32 & 79.30\textsubscript{\scriptsize{(6.38)}} & 81.37\textsubscript{\scriptsize{(6.05)}} \\ \bottomrule
\end{tabular}
\label{table_cifar}
\end{table}

\subsubsection{FEMNIST}
Table \ref{table_emnist} shows results on FEMNIST with 10 and 20 clients. The normalization approach yields substantial improvements, with up to 9\% accuracy gain for FedNova with 20 clients. SGDM consistently outperforms other methods, suggesting the momentum term's effectiveness in handling the natural heterogeneity of handwritten characters. The performance improvement is more significant with 20 clients, indicating our method's scalability to larger client pools. FedNova shows notable improvement when normalized, particularly in the 20-client scenario, demonstrating its effectiveness in managing varying local update frequencies across clients.


\begin{table}[t]
\setlength{\tabcolsep}{4pt}
\centering
\caption{Performance comparison on FEMNIST with varying number of clients. The numbers in brackets indicate the increase in accuracy from the original to the normalized approach.}
\begin{tabular}{ccccc}
\toprule
\textbf{Approach} & \multicolumn{2}{c}{\textbf{Original}} & \multicolumn{2}{c}{\textbf{Normalized}} \\ 
 & \textbf{10 Clients} & \textbf{20 Clients} & \textbf{10 Clients} & \textbf{20 Clients} \\ \midrule
\textbf{FedAvg}   & 61.06 & 67.89 & 67.28\textsubscript{\scriptsize{(6.22)}} & 75.18\textsubscript{\scriptsize{(7.29)}} \\ 
\textbf{FedProx}  & 64.08 & 69.37 & 68.40\textsubscript{\scriptsize{(4.32)}} & 74.44\textsubscript{\scriptsize{(5.07)}} \\ 
\textbf{SCAFFOLD} & 65.08 & 71.17 & 70.08\textsubscript{\scriptsize{(5.00)}} & 76.20\textsubscript{\scriptsize{(5.03)}} \\ 
\textbf{SGDM}     & 66.89 & 71.18 & 71.98\textsubscript{\scriptsize{(5.09)}} & 77.14\textsubscript{\scriptsize{(5.96)}} \\ 
\textbf{FedNova}  & 62.26 & 66.01 & 71.12\textsubscript{\scriptsize{(6.86)}} & 75.64\textsubscript{\scriptsize{(9.63)}} \\ 
\textbf{FedBABU}  & 65.50 & 67.86 & 72.32\textsubscript{\scriptsize{(6.82)}} & 76.64\textsubscript{\scriptsize{(8.78)}} \\ \bottomrule
\end{tabular}
\label{table_emnist}
\end{table}

\begin{figure}
    \centering
    \includegraphics[width=0.9\linewidth]{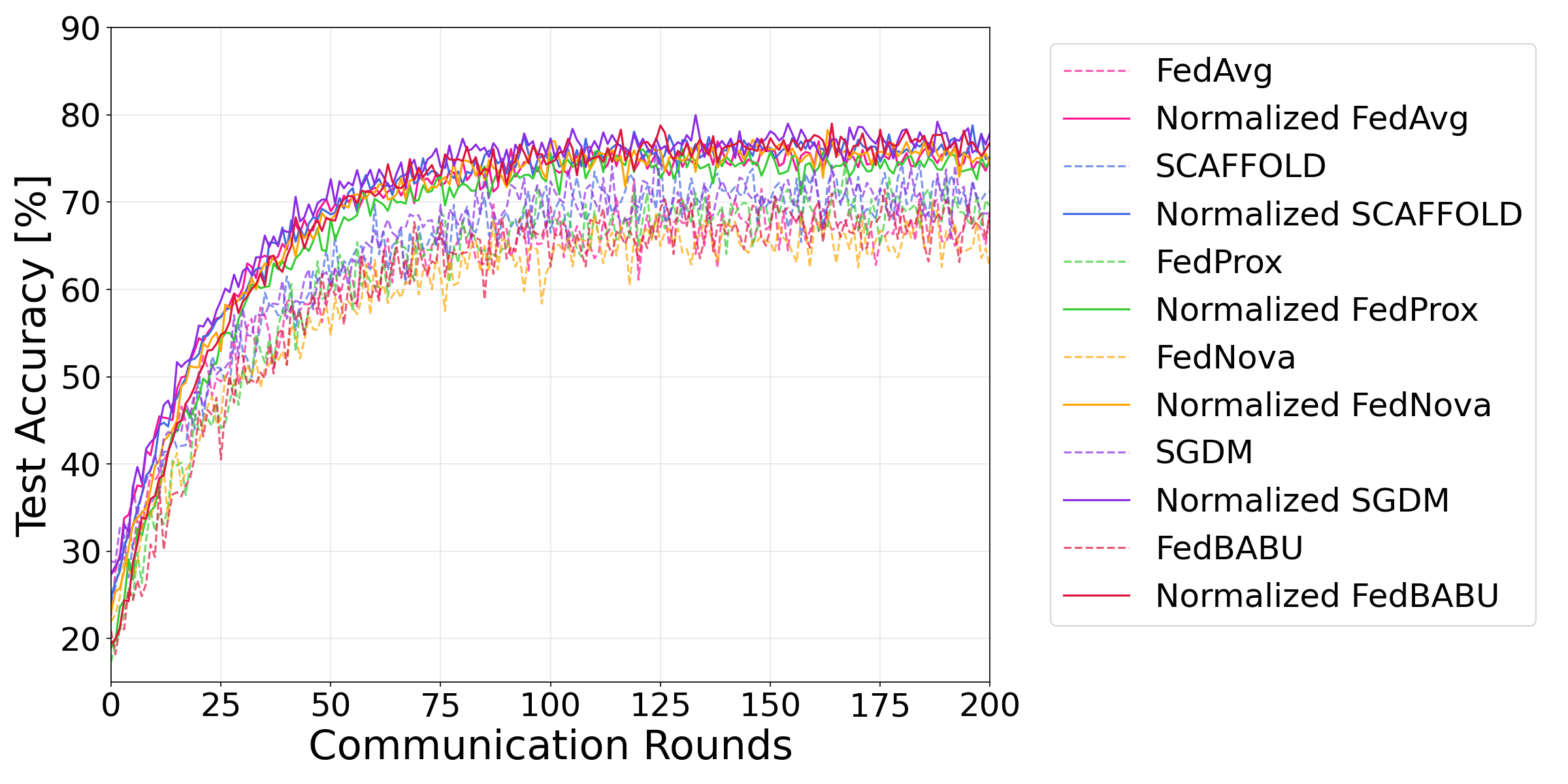}
    \caption{Convergence analysis of FL schemes on FEMNIST. Normalized approaches (solid lines) demonstrate consistently superior performance compared to original methods (dashed lines).}
    \label{con}
\end{figure}

Figure \ref{con} illustrates the convergence behavior of different methods on FEMNIST. The normalized approaches consistently achieve higher accuracy and faster convergence, particularly in the early stages of training. This acceleration in convergence is crucial for reducing communication rounds in practical FL deployments, where network efficiency is a key concern.

\subsubsection{CINIC-10}
Table \ref{table_cinic} presents results on CINIC-10 with 100 clients and $\alpha = 0.2, 0.4, 0.6,$ and $0.8$. The normalization approach consistently outperforms original methods across all $\alpha$ values, with improvements ranging from 2\% to 7\%. The performance gap between normalized and original methods narrows as $\alpha$ increases, indicating our method's particular effectiveness in highly heterogeneous settings. SGDM with normalization achieves the highest accuracy (85.04\% for $\alpha = 0.8$), suggesting a synergistic effect between momentum-based optimization and our contribution-based normalization. Interestingly, FedProx shows significant improvement when normalized, especially at higher $\alpha$ values, indicating that the proximal term combined with our normalization approach effectively manages client drift in less heterogeneous scenarios.

\begin{table}[t]
\centering
\setlength{\tabcolsep}{1pt}
\caption{Comparison of experimental results for FL schemes with and without normalization on the CINIC-10 Dataset. 'Org' stands for Original and 'Norm' stands for Normalization.}
\begin{tabular}{cccccccc}
\toprule
\textbf{$\alpha$} & \textbf{Approach} & \textbf{FedAvg} & \textbf{FedProx} & \textbf{SCAFFOLD} & \textbf{SGDM} & \textbf{FedNova} & \textbf{FedBABU} \\
\midrule
\multirow{2}{*}{$0.2$} & Org & 71.16 & 72.54 & 71.38 & 71.68 & 70.44 & 71.96 \\
& Norm & 77.24 & 76.54 & 78.52 & 78.30 & 77.74 & 78.06 \\
\midrule
\multirow{2}{*}{$0.4$} & Org & 73.74 & 74.92 & 73.92 & 76.12 & 74.46 & 74.14 \\
& Norm & 78.08 & 79.46 & 77.68 & 82.48 & 79.12 & 79.86 \\
\midrule
\multirow{2}{*}{$0.6$} & Org & 76.00 & 77.46 & 76.62 & 78.36 & 77.14 & 77.78 \\
& Norm & 78.88 & 81.02 & 80.14 & 84.06 & 81.62 & 81.22 \\
\midrule
\multirow{2}{*}{$0.8$} & Org & 77.64 & 78.38 & 77.20 & 79.52 & 78.02 & 78.20 \\
& Norm & 79.54 & 81.96 & 80.10 & 85.04 & 82.22 & 82.10 \\
\bottomrule
\end{tabular}
\label{table_cinic}
\end{table}



\subsection{Discussion}
Our experimental results demonstrate the efficacy of the normalization approach across diverse data settings and baseline FL methods. The normalization approach enhances accuracy across all tested datasets, FL methods, and heterogeneity levels, showing particular effectiveness in highly heterogeneous settings (low $\alpha$ values). This improvement is crucial for real-world FL applications where data distributions across clients are often highly skewed. The consistent superior performance of SGDM with normalization suggests a complementary relationship between momentum-based optimization and our contribution-based normalization, potentially opening avenues for further research into hybrid optimization strategies in FL.

The approach's effectiveness scales well with increasing client numbers, as evidenced by the FEMNIST results, indicating its suitability for large-scale FL deployments. Normalized methods consistently achieve faster convergence, particularly in early training stages, potentially reducing communication rounds in practical FL deployments. This acceleration in convergence is especially valuable in bandwidth-constrained or high-latency network environments, where minimizing communication overhead is critical.

The performance improvements across different methods provide insights into the interaction between our approach and existing FL techniques. For instance, the significant improvement in FedNova's performance when normalized suggests that our approach effectively complements methods designed to handle heterogeneity. Similarly, the enhanced performance of FedProx with normalization indicates that our method can work synergistically with proximal term regularization to manage client drift. These findings underscore the proposed approach's potential in enhancing FL performance across diverse real-world scenarios characterized by data heterogeneity.

\section{Conclusion}
This paper introduces a novel approach to address statistical heterogeneity in FL through mean latent representations from locally trained models. Our method normalizes client contributions during aggregation, effectively mitigating the challenges of non-IID data distributions across diverse clients. The proposed normalization technique shows significant advantages such as low computational overhead, broad applicability, and seamless integration with various FL frameworks. This research opens up several promising avenues for future work, including the extension of the contribution factor for optimized client selection and malicious client detection, exploration of adaptive normalization techniques and integration with privacy-preserving mechanisms. As FL continues to gain prominence in privacy-preserving distributed machine learning, our approach represents a significant advancement in addressing the challenges of statistical heterogeneity, paving the way for more robust and efficient FL systems across diverse real-world applications.

\bibliographystyle{IEEEtranS}
\bibliography{main}

\end{document}